\pdfoutput=1
\documentclass[11pt]{article}

\usepackage[final]{acl}

\PassOptionsToPackage{table}{xcolor}
\PassOptionsToPackage{a4paper,top=4.2cm,bottom=4.2cm,left=3.5cm,right=3.5cm}{geometry}

\usepackage{times}
\usepackage{latexsym}

\usepackage[T1]{fontenc}
\usepackage[utf8]{inputenc}

\usepackage{microtype}

\usepackage{inconsolata}

\usepackage{graphicx}
\usepackage{listings}
\usepackage{xcolor}

\usepackage{geometry} % for setting page size and margins
\usepackage{amsmath}   % for equations and mathematics
\usepackage{booktabs}  % For professional-looking tables
\usepackage{array}
\usepackage{siunitx}   % For aligning decimal points in numbers
\usepackage{listings}    % for code listings
\usepackage{hyperref}  % for hyperlinks
\usepackage{multirow}

\newcolumntype{C}{>{\centering\arraybackslash}p{1.5cm}} % Example of a centered paragraph column type
\newcolumntype{P}[1]{>{\centering\arraybackslash}p{#1}} % Define P column type for a paragraph with specified width

\title{Rapid Biomedical Research Classification: \\The Pandemic PACT Advanced Categorisation Engine}

\author{
  Omid Rohanian$^{1,2}$,
  Mohammadmahdi Nouriborji$^{1,7}$,
  Olena Seminog $^{3}$,\\
  \textbf{Rodrigo Furst}$^{3}$,
  \textbf{Thomas Mendy}$^{3}$,
  \textbf{Shanthi Levanita}$^{3}$,
  \textbf{Zaharat Kadri-Alabi}$^{3}$,\\
  \textbf{Nusrat Jabin}$^{3}$,
  \textbf{Daniela Toale}$^{3}$,
  \textbf{Georgina Humphreys}$^{5}$,\\
  \textbf{Emilia Antonio}$^{3}$,
  \textbf{Adrian Bucher}$^{6}$,
  \textbf{Alice Norton}$^{3}$,
  \textbf{David A. Clifton}$^{2,4}$\\ \\
  $^1$NLPie Research, Oxford, UK \\
  $^2$Department of Engineering Science, University of Oxford, Oxford, UK \\
  $^3$Pandemic Sciences Institute, University of Oxford, Oxford, England, UK \\
  $^4$Oxford-Suzhou Centre for Advanced Research, Suzhou, China \\
  $^5$Green Templeton College, University of Oxford, Oxford, England, UK\\
  $^6$UK Collaborative on Development Research, London, UK\\
  $^7$Sharif University of Technology, Tehran, Iran \\ \\
  \texttt{omid.rohanian@eng.ox.ac.uk}
 }

\begin{document}
\maketitle
\vspace{0.5cm}

\begin{abstract}
This paper introduces the Pandemic PACT Advanced Categorisation Engine (PPACE) along with its associated dataset. PPACE is a fine-tuned model developed to automatically classify research abstracts from funded biomedical projects according to WHO-aligned research priorities. This task is crucial for monitoring research trends and identifying gaps in global health preparedness and response. Our approach builds on human-annotated projects, which are allocated one or more categories from a predefined list. A large language model is then used to generate `rationales' explaining the reasoning behind these annotations. This augmented data, comprising expert annotations and rationales, is subsequently used to fine-tune a smaller, more efficient model. Developed as part of the Pandemic PACT project, which aims to track and analyse research funding and clinical evidence for a wide range of diseases with outbreak potential, PPACE supports informed decision-making by research funders, policymakers, and independent researchers. We introduce and release both the trained model\footnote{\url{https://huggingface.co/nlpie/ppace-v1.0}} and the instruction-based dataset used for its training\footnote{\url{https://huggingface.co/datasets/nlpie/pandemic_pact}}. Our evaluation shows that PPACE significantly outperforms its baselines. The release of PPACE and its associated dataset offers valuable resources for researchers in multilabel biomedical document classification and supports advancements in aligning biomedical research with key global health priorities.
\end{abstract}

\section{Introduction}
\label{intro}

The surveillance and monitoring of emerging and re-emerging pathogens are vital to global health security. Infectious diseases with the potential to cause pandemics represent a significant threat to public health, economies, and societies worldwide. Efficiently tracking these threats and coordinating research efforts is essential to mitigate their impact. Traditional approaches to research funding and coordination during health crises, such as the COVID-19 pandemic, often exhibit limitations including slow activation of research projects, duplication of efforts, and fragmented funding landscapes. These issues highlight the need for improved systems to manage and analyse research activities \citep{carroll2021preventing, mclean2022fragmented}.

The rapid identification and response to infectious disease threats require a well-coordinated and systematic approach to research funding and project tracking. Accurate categorisation and analysis of research projects are crucial for identifying trends, understanding research gaps, and ensuring that resources are allocated effectively. This task is inherently complex, given the diverse and interdisciplinary nature of biomedical research \citep{Seminog2024}.

Artificial intelligence (AI), particularly large language models (LLMs), offers a promising solution to enhance the efficiency and accuracy of research categorisation. LLMs can be fine-tuned to assist in the classification of research abstracts, providing valuable support to human annotators. These models not only streamline the categorisation process but also generate rationales for their decisions, adding a layer of interpretability and transparency that is essential for gaining the trust of researchers and policymakers\footnote{LLMs are typically seen as complex and opaque, and their interpretability is a nuanced, ongoing research topic \citep{luo2024understanding}. Here, we use ``interpretable'' to mean that medical practitioners find the process more engaging and understandable than with encoder-only models, as LLMs can explain their reasoning in human language.}.

This paper introduces the Pandemic PACT Advanced Categorisation Engine (PPACE), a fine-tuned LLM designed to classify biomedical research abstracts according to WHO-aligned research priorities. PPACE leverages human-annotated data and employs generative AI to produce rationales for each classification. By automating the categorisation process, PPACE aims to enhance the monitoring of research trends and the identification of critical gaps in global health preparedness and response.

In the remainder of the paper, we first provide an overview of the literature on the use of LLMs in biomedical document classification (Section \ref{background}). Next, we introduce the Pandemic PACT project which this work builds on, and describe the details of the dataset and the annotation procedure involved in its creation (Section \ref{dataset}). Section \ref{method} will describe the methodology in finetuning the PPACE model, and finally, in Section \ref{results}, we present the results and conclude the paper. The contributions of this work are as follows:

\begin{enumerate}
    \item We contribute to the task of biomedical document classification by publicly releasing a carefully annotated dataset of research projects (each project containing a title and a PubMed-style abstract) gathered as part of the Pandemic PACT project and further preprocessed to include rationales generated by a 70B LLM. The augmented dataset is formatted as an instruction-based dataset and can be used to train similar models by the research community.
    \item We fine-tune and publicly release an 8B model trained on the aforementioned dataset and make the model weights available publicly.
    \item We perform a range of analyses on the dataset to shed light on the complexities of the data and run a number of evaluations to ensure that the model outperforms the baseline.
\end{enumerate}

\section{Biomedical Document Classification and the Use of LLMs}
\label{background}

Biomedical document classification is an active area of research that has attracted considerable attention in recent years \citep{laza2011evaluating}. The PubMed 200k RCT dataset, for instance, focuses on classifying different sections of randomized controlled trial abstracts into categories like objectives, methods, results, and conclusions \citep{dernoncourt2017pubmed}. Another notable task in this area is the Hallmarks of Cancer (HoC), which presents a multi-label classification challenge and aims to identify key cancer-related research themes from PubMed abstracts \citep{baker2015automatic}. The LitCovid dataset \citep{chenlitcovid2020, jimenez-gutierrez-etal-2020-document}, which comprises over 30,000 COVID-19-related articles, each annotated with one or more topics relevant to the pandemic, is another major biomedical document classification benchmark that has been studied in the literature. Automated topic annotation tasks like this can significantly reduce the manual curation burden during pandemics, and the present work can be considered a more generalized effort to automate the classification of biomedical literature into research themes of interest. Additional datasets relevant to biomedical document classification include the BioCreative Corpus III \citep[BC3,][]{arighi2011biocreative} and TREC \citep{hersh2006trec}. \citet{behera2019performance} provides an overview of this task and the various deep learning algorithms to address it.

Large Language Models (LLMs) have become ubiquitous in various text processing and classification tasks, including document classification. Their ability to handle a wide range of text-related tasks makes them particularly appealing for numerous applications. LLMs can be instructed to perform specific tasks via few-shot examples or through fine-tuning with detailed instructions. For biomedical researchers, generative LLMs are especially valuable because they can be interfaced with using natural human language, facilitating more intuitive and effective interactions.

SciFive \citep{phan2021scifive} is a domain-specific T5 model that has been pretrained to address a number of biomedical tasks, including document classification. \citet{rohanian2023exploring} is the first attempt to use generative language models to address classical biomedical text processing tasks like HoC via instruction tuning. \citet{chen2023large} and \citet{tian2024opportunities} have studied the use of LLMs in a number of biomedical text processing tasks, including document classification, although the focus is mostly on closed-source frozen models like GPT-4.

Various techniques are observed in the literature regarding the use of LLMs when addressing this task. Several studies use parameter-efficient fine-tuning methods \citep{hu2021lora, taylor2024efficiency, jiang2024mora} which have become very prevalent due to the ease of use and faster training time they offer. Our work not only employs instruction tuning and LoRA as a parameter-efficient fine-tuning technique, but also draws inspiration from \citet{hsieh2023distilling} in that it utilises `rationales' generated by a larger model to augment the labelled dataset and then fine-tunes a smaller one trained on the expanded dataset.

\section{Dataset Overview and Sources}
\label{dataset}

\subsection{About Pandemic PACT}
The Pandemic Preparedness Analytical Capacity and Funding Tracking Programme (Pandemic PACT) operates under the auspices of the Global Research Collaboration for Infectious Disease Preparedness \citep[GloPID-R,][]{norton2020preparing}\footnote{\url{https://www.glopid-r.org/}} at the University of Oxford's Pandemic Sciences Institute. This initiative aims to enhance global response capabilities by tracking and analysing research funding for diseases with pandemic potential and other significant public health threats. By aligning with the WHO priority diseases, Pandemic PACT focuses on dynamic data collection and rigorous analysis to inform critical policy and funding decisions across the health system and public health domains \citep{norton2024improving,Seminog2024}.

\subsection{The Pandemic PACT Funding  Tracker}
The Pandemic PACT Funding Tracker is an integral component of this initiative, collecting detailed information on research grants from GloPID-R and UKCDR\footnote{The UK Collaborative on Development Research (UKCDR) coordinates development research funding in the UK to optimise effectiveness and strategic alignment. More information is available at \url{https://www.ukcdr.org.uk}.
} members since January 2020 and has since expanded to include a much broader set of international funding bodies. This tool maps the alignment of funding to critical research categories and priorities, displaying the data through an interactive dashboard that visualises funding trends and evidence gaps. The database includes diseases listed on the WHO R\&D Blueprint priority list, such as pandemic influenza, mpox, and plague, among others. This comprehensive and evolving tool not only aids in real-time decision-making but also provides downloadable data for broader analysis, accessible via the official Pandemic PACT website at \url{http://www.pandemicpact.org/}.

\subsection{Annotation Procedure}
\label{annotation}

The Pandemic PACT database expands upon the previous database co-developed by UKCDR and GloPID-R as part of the COVID-19 Research Coordination and Learning initiative (COVID CIRCLE1). Pandemic PACT funding data on other diseases and additional COVID-19 research projects are collected either through direct data provision by funders (using a standardised template) or by scraping funder websites. The scraping process is based on search terms including disease-specific keywords, acronyms, virus, and virus family names. For the detailed search protocol, inclusion criteria, and transformation of the COVID CIRCLE data into the new standardised schema, see \citet{Seminog2024}.
Only grants that include a minimum level of essential information are included, such as grant award or start date, publication date, funder name, grant ID or another form of identifier, and grant title. The data encompass funding information from January 2020 onwards for the relevant diseases. 

While all Pandemic PACT search terms used are in English, it does not exclude grants in other languages. If the search returns any relevant grants in foreign languages, their title and abstracts are translated using Google Translate and then included in the database. All collected data is stored in its original format as retrieved from the funding source, with basic data cleaning procedures performed to remove special characters from data in textual format. 

All collected data are reviewed by a team of trained researchers from broad public health backgrounds to determine their relevance, classified against a research categorisation framework developed under Pandemic PACT, and assigned other relevant tags using manual annotation. The number of team members has varied over time, starting with three, increasing to ten before the launch of the Pandemic PACT tracker, then decreasing to six, and currently stabilising at four. The team size is subject to change based on project needs. Over the first months of the project, Pandemic PACT developed a standard approach to training and preparing new members of the data collection team through a series of training steps. First, they were exposed to tutorials of training material and videos that explained how to interpret data and submit contributions through the online interface. After that, data coders were expected to attend a weekly all-contributor meeting, at which point they started being included in the regular coding allocation. These meetings were used for expanding comprehension of the coding schema and processes, facilitating a collective consensus on interpretations of codes, and effectively probing coding disagreements. 

\begin{table}[ht]
\centering
\small
\begin{tabular}{@{}lc@{}}
\toprule
\textbf{Dataset} & \textbf{Number of Projects} \\ 
\midrule
Training Set & 5142 \\ 
Test Set & 1450 \\ 
\bottomrule
\end{tabular}
\caption{Composition of the dataset used for training and testing the classification model.}
\label{tab:dataset-size}
\end{table}

After data is entered, they are marked as `unverified' in the back-end database portal used by the Pandemic PACT if any issues arise or if the coder hesitates on how to code them. This flags them for the review process. Conversely, entries are marked as `complete' if no concerns are raised. To ensure data reliability, Pandemic PACT mandates peer review of all new data by at least two annotators, ensuring each grant undergoes scrutiny and confirmation by an independent coder. In cases of inter-annotator disagreement, discussions are held to reach joint decisions. Alternatively, judgments from a designated coder, such as the Principal Investigator or a more experienced researcher, take precedence over others. Going forward, Pandemic PACT plans to implement a comprehensive approach where initial coding is performed by an LLM, followed by manual verification and final annotation.

\begin{table*}[ht]
\centering
\small
\begin{tabular}{@{}cl@{}}
\toprule
\textbf{Category Number} & \textbf{Research Category} \\ 
\midrule
1 & Pathogen: Natural History, Transmission, and Diagnostics \\ 
2 & Animal and Environmental Research \& Research on Diseases Vectors \\ 
3 & Epidemiological Studies \\ 
4 & Clinical Characterisation and Management in Humans \\ 
5 & Infection Prevention and Control \\ 
6 & Therapeutics Research, Development, and Implementation \\ 
7 & Vaccines Research, Development, and Implementation \\ 
8 & Research to Inform Ethical Issues \\ 
9 & Policies for Public Health, Disease Control, and Community Resilience \\ 
10 & Secondary Impacts of Disease, Response, and Control Measures \\ 
11 & Health Systems Research \\ 
12 & Capacity Strengthening \\ 
\bottomrule
\end{tabular}
\caption{The full list of research categories used to annotate the dataset. }
\label{tab:research-categories}
\end{table*}

\subsection{Dataset Description}
Our study employs a carefully selected sample from the Pandemic PACT database. Each row in the data represents a funded research project and includes a title and an abstract which provides a concise description of each project's aims, methods, and potential impacts. The data is randomly divided into an approximate 80/20 split with the number of rows shown in Table \ref{tab:dataset-size}.

To gain insights into the training set, we analysed the lengths of the project titles and abstracts as well as the distribution of research categories. The statistics for the lengths of project titles and abstracts are presented in Table~\ref{tab:length-stats}. During finetuning, to keep the computation manageable, the abstract length is capped at 512 tokens. The numbers in the table  reflect the lengths as seen in the dataset before this truncation is applied.

\begin{table}[ht]
\centering
\small
\begin{tabular}{@{}llcc@{}}
\toprule
& \textbf{Measure} & \textbf{Title} & \textbf{Abstract} \\ 
\midrule
\multirow{2}{*}{\textbf{Characters}} & Average Length & 98.24 & 1940.37 \\ 
& Max Length & 850 & 6817 \\ 
\midrule
\multirow{2}{*}{\textbf{Words}} & Average Length & 13.10 & 279.72 \\ 
& Max Length & 133 & 1036 \\ 
\bottomrule
\end{tabular}
\caption{Statistics of project titles and abstracts in the training set. The measure of words is an approximation based on space separation.}
\label{tab:length-stats}
\end{table}

\begin{figure}[ht]
\centering
\includegraphics[width=0.95\columnwidth]{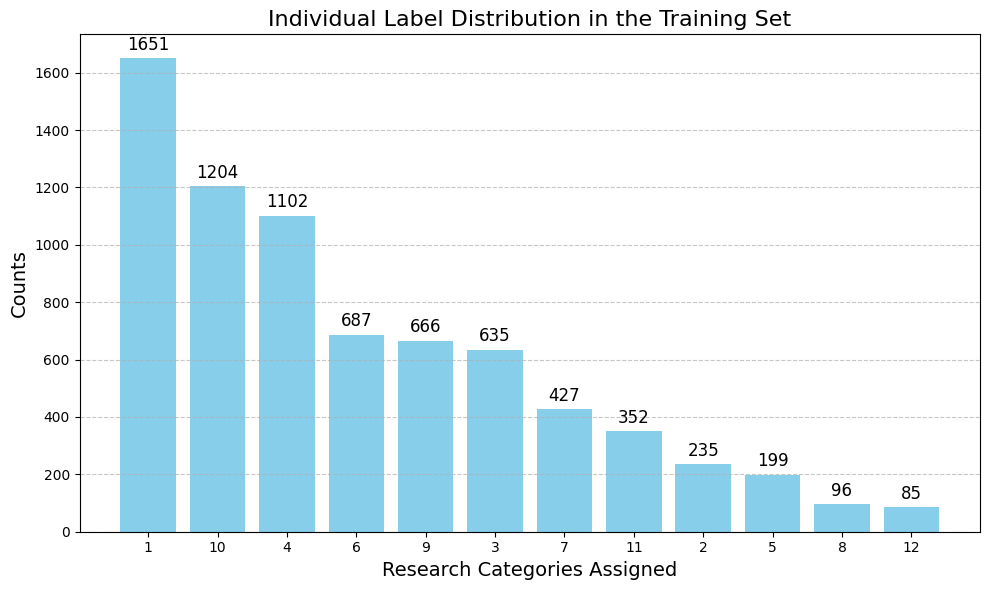}
\caption{Individual Label Distribution in the Training Set.}
\label{fig:individual-label-dist}
\end{figure}

The distribution of individual research categories (see Table \ref{tab:research-categories}) assigned to the projects in the training set is depicted in Figure~\ref{fig:individual-label-dist}. This figure shows that the most frequent research categories are Pathogen: Natural History, Transmission, and Diagnostics (Category 1), Secondary Impacts of Disease, Response, and Control Measures (Category 10), and Clinical Characterisation and Management in Humans (Category 4). The least frequent categories are Research to Inform Ethical Issues (Category 8), Capacity Strengthening (Category 12), and Infection Prevention and Control (Category 5). These categories are expected to pose more challenges for the model due to the fewer number of labels.

\begin{figure}[ht]
\centering
\includegraphics[width=0.95\columnwidth]{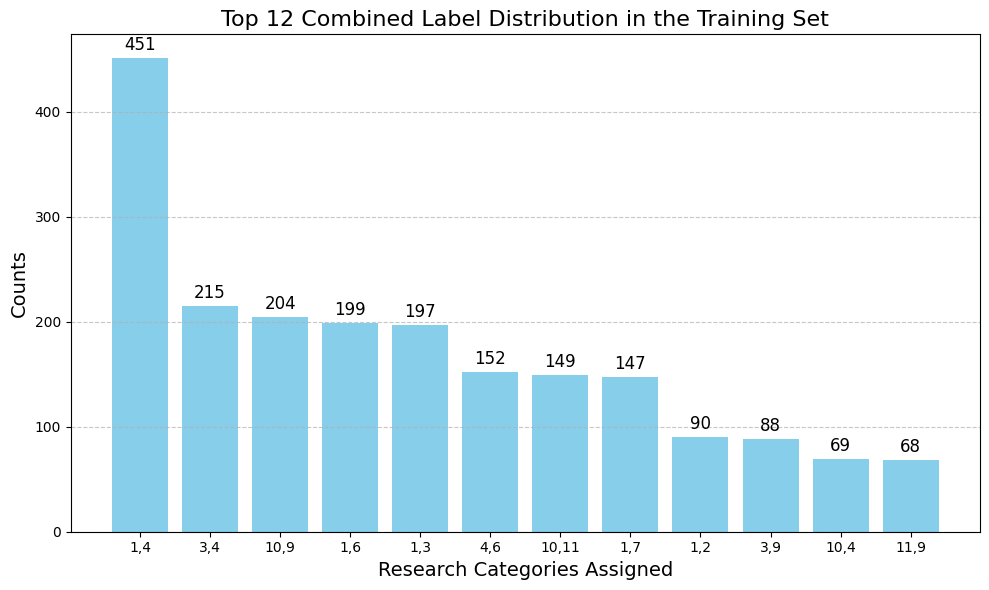}
\caption{Top 12 Combined Label Distribution in the Training Set.}
\label{fig:combined-label-dist}
\end{figure}

\subsection{Combined Label Distributions}
We also examined the combined label distributions to understand the most common combinations of research categories assigned to the projects. Figure~\ref{fig:combined-label-dist} shows the top 12 most frequent label clusters, indicating that combinations such as Pathogen \& Clinical Characterisation in Humans (1, 4), and Policies for Public Health \& Secondary Impacts of Disease (9, 10) are prevalent. Table~\ref{tab:research-categories} provides the mapping of category numbers to their respective research categories.

\subsection{Label Correlations}
Understanding the correlations between different research categories provides insights into interdisciplinary research trends visible in the training set. These correlations highlight how different fields of study intersect, helping us identify areas where models might struggle or easily pick up patterns.

Apart from the significant correlations mentioned in Figure~\ref{fig:combined-label-dist}, we also found notable intersections between Epidemiological Studies (Category 3) and Clinical Characterisation in Humans (Category 4), and between Pathogen (Category 1) and Therapeutics Research (Category 6). A properly trained model should be able to detect these patterns while also recognising instances where these correlations do not hold.

The conditional probabilities for the top five most frequent pairs of research categories are shown in Table \ref{tab:conditional-probabilities}. For example, the highest conditional probability for the combination (1, 3) is for label 4 at 0.39, and for (1, 4), the highest is label 3 at 0.17. These findings suggest that certain third-label correlations exist, but they are not overwhelmingly strong.

\begin{table}[ht]
\centering
\small
\begin{tabular}{@{}ccc@{}}
\toprule
\textbf{Combination} & \textbf{Top Conditional Probability} & \textbf{Probability} \\ 
\midrule
\{1, 4\} & P(3 | \{1, 4\}) & 0.17 \\ 
\{3, 4\} & P(1 | \{3, 4\}) & 0.35 \\ 
\{10, 9\} & P(11 | \{10, 9\}) & 0.12 \\ 
\{1, 6\} & P(4 | \{1, 6\}) & 0.27 \\ 
\{1, 3\} & P(4 | \{1, 3\}) & 0.39 \\ 
\bottomrule
\end{tabular}
\caption{Top conditional probabilities for the most frequent pairs of research categories in the training set.}
\label{tab:conditional-probabilities}
\end{table}

\begin{figure*}[ht]
\centering
\includegraphics[width=0.65\textwidth]{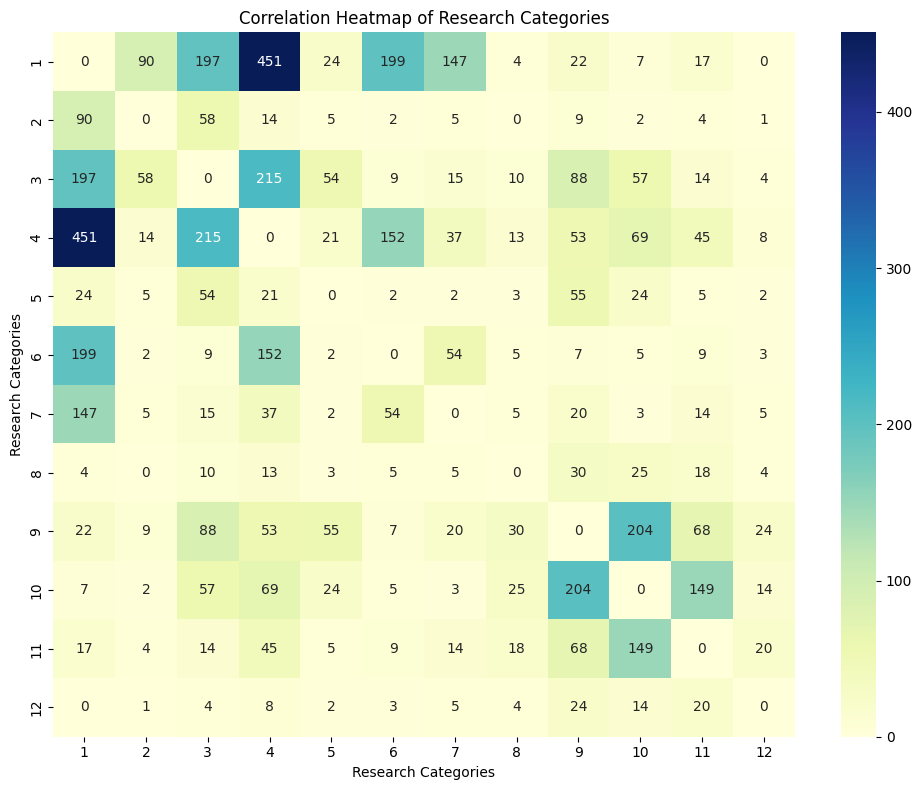}
\caption{Correlation Heatmap of Research Categories in the Training Set.}
\label{fig:label-correlations-heatmap}
\end{figure*}

The heatmap in Figure~\ref{fig:label-correlations-heatmap} provides a visual representation of the strength of correlations between different research categories.

\section{Methodology}
\label{method}

In this work, we initially use a manually labelled dataset to generate `rationales' for the labels using a Llama-3 70B model\footnote{\url{https://huggingface.co/meta-llama/Meta-Llama-3-70B-Instruct}}. These rationales explain why each category label is chosen, pinpointing the reason by referencing the abstract. We add these rationales to the labels, constructing an expanded dataset that includes both the prompt and the rationales. For details on the prompt template used, refer to Appendix Section \ref{sec:appendix_prompt}.

We subsequently explore adapting a smaller model to the classification task using this dataset. The fine-tuned model is expected not only to predict the labels but also to explain its reasoning. Based on insights from \citet{mccoy2023embers} regarding the limitations of autoregressive language models, we structured the prompt such that the rationales are provided first before the model determines the category labels. This approach ensures that errors from irrelevant categories are not propagated back into the model's outputs. Given the substantial computing resources required to train the full weights of the 8B model, we explored the use of efficient fine-tuning via LoRA (Section \ref{subsec:lora}).

We have chosen to restrict the experiments and the baselines to a single decoder-only transformer. As of this writing, an LLM with around 10 billion parameters is considered relatively small but can be performant enough to rival state-of-the-art. A representative LLM with a good starting performance on this complex task provides us with a foundation to improve upon while avoiding potential saturation. Additionally, a very large model like Llama-3 70B would be impractical for independent researchers to fine-tune or utilise on less powerful machines. Section \ref{subsec:adju} details how this LLM was chosen. 

\subsection{Adjudicating between Outputs of Candidate LLMs}
\label{subsec:adju}

In order to fine-tune a smaller model using the augmented dataset of human annotations and rationales generated by the LLama 3-70B, we evaluated several candidate models. The best-performing ones were the Mixtral-8x7B Instruct\footnote{\url{https://huggingface.co/mistralai/Mixtral-8x7B-Instruct-v0.1}} and the Meta-Llama-3-8B-Instruct\footnote{\url{https://huggingface.co/meta-llama/Meta-Llama-3-8B-Instruct}}. We empirically found that the outputs produced by these models were significantly more relevant compared to other models of similar or smaller size such as Phi3.

To determine which model to fine-tune for developing PPACE, we conducted a detailed comparison. We randomly selected 10 projects from the dataset and used a few-shot learning template (Section \ref{sec:appendix_prompt}) to obtain inferences from both Mixtral and Llama3 models, utilising them as frozen language models. Each model's output was then recorded and passed to GPT-4o for adjudication. GPT-4o was tasked with evaluating the responses, comparing them to the available human judgments, and providing a verdict favouring one model output over the other or declaring a tie.

This adjudication process involved a thorough analysis of each model's outputs against human labels and was further verified by our annotators. The results showed that both models performed well, but Llama-3 was deemed the better model by a small margin. One of the key advantages of Llama-3 was that extracting the output labels from the generated text was significantly easier compared to Mixtral, which occasionally deviated from the specified format, complicating the extraction process. Additionally, Mixtral has 46.7B total parameters but only uses 12.9B parameters per token. In other words, Llama-3 also has the advantage of being smaller and more performant for this task. This consistency and ease of label extraction made Llama-3 the preferred choice for fine-tuning PPACE.

\subsection{Low-Rank Adaptation (LoRA)}
\label{subsec:lora}

Low-Rank Adaptation \citep[LoRA,][]{hu2021lora} is a parameter-efficient approach for adapting large pre-trained models without modifying the original weights. This method can be particularly beneficial for maintaining memory efficiency and reducing computational overhead. LoRA introduces two small matrices, \( \mathbf{A} \) and \( \mathbf{B} \), into the transformer architecture, which project the high-dimensional parameter space into a lower-dimensional space and back. The original transformation in a transformer layer, typically a matrix multiplication involving a weight matrix \( \mathbf{W} \), is modified as follows:

\begin{equation}
    \mathbf{W}' = \mathbf{W} + \frac{r}{\alpha}\mathbf{A}\mathbf{B}
\end{equation}

Here, \( \mathbf{W} \) is the original weight matrix of the transformer, and \( \mathbf{W}' \) is the adapted weight matrix. The matrices \( \mathbf{A} \) and \( \mathbf{B} \) are of dimensions \( d \times r \) and \( r \times d \) respectively, where \( d \) is the dimensionality of \( \mathbf{W} \) and \( r \) is much smaller than \( d \), finally $\alpha$ is a hyperparameter for adjusting the learning rate for the trainable weights. This low-rank structure ensures that the number of additional parameters introduced by \( \mathbf{A} \) and \( \mathbf{B} \) is significantly lower than the number of parameters in \( \mathbf{W} \), leading to substantial savings in terms of memory and computational resources.

The low-rank projection effectively captures the essential transformations needed for task-specific adaptation while preserving the original model's capabilities. This approach is particularly advantageous when computational resources are limited or when the adaptability needs to be achieved with minimal disturbance to the original model structure, as often required in real-world applications where both efficiency and performance are critical. In practical applications, LoRA has shown to enable effective fine-tuning of large models on specific tasks without the need for extensive retraining of the original parameters.

\subsection{Training Strategy and Hyperparameters}
We used Supervised Finetuning (SFT) to adapt the Llama-3 8B model to the classification task. The model was trained for $2$ epochs on the training samples using $8$ A100 GPUs, with a batch size of $1$ per GPU and $4$ gradient accumulation steps. The LoRA modules were placed in all trainable layers of the self-attention and MLP layers of the Llama model.

During initial experiments, we found that focusing the loss calculation on the completion tokens (related to the explanations and categories) and ignoring the loss for the prompt tokens improved the model's performance. This approach was more effective than using the autoregressive language modelling loss on all tokens, including the prompt.

To maximise model performance, we used a beam search decoding strategy with $4$ beams for the smaller models, which appeared to markedly improve generation quality and output structure. However, due to computational constraints, beam search was not employed for the $70$B version.

The hyperparameters used for training the model in our experiments are listed in Table \ref{tab:hyperparameters}.

\begin{table}[ht!]
    \centering
    \caption{The hyperparameters used for training the model} \vspace{1pt}
    \begin{tabular}{p{5cm}P{2cm}} % Adjusted column width
        \toprule 
        \textbf{Hyperparameter} & \textbf{Value} \\
        \cmidrule(lr){1-2} % Corrected cmidrule command
        Total Batch Size & $8$ \\
        Gradient Accumulation Steps & $4$ \\
        Learning Rate & $2e$-$4$ \\
        LR Scheduler & Linear \\
        Epochs & $2$ \\
        LoRA rank & $128$ \\
        LoRA $\alpha$ & $256$ \\
        LoRA Dropout & $0.05$ \\
        \bottomrule
    \end{tabular}
    \label{tab:hyperparameters}
\end{table}

% Define the custom column types with unique names
\newcolumntype{Y}[1]{>{\centering\arraybackslash}p{#1}} % centered paragraph column
\newcolumntype{Z}[1]{>{\centering\arraybackslash}p{#1}} % centered paragraph column

\begin{table*}[h!]
    \centering
    \caption{The results of different models on the test set} \vspace{10pt} % Add a title above the table 
    \begin{tabular}{Y{4cm}Z{3cm}Z{3cm}Z{3cm}} % Use the new Y and Z types
        \toprule 
        \textbf{Model} & \textbf{Precision} & \textbf{Recall} & \textbf{F1}  \\
        & \texttt{Macro/Micro} & \texttt{Macro/Micro} & \texttt{Macro/Micro} \\
        \midrule
        Llama3-8b & 0.2710/0.2631 & 0.5812/0.6042 & 0.3293/0.3666\\
        Llama3-70b & 0.3163/0.3524 & \textbf{0.6320}/0.6515 & 0.3898/0.4574\\
        PPACE (ours) & \textbf{0.6927}/\textbf{0.7497} & 0.5625/\textbf{0.7113} & \textbf{0.5914}/\textbf{0.7300}\\
        \bottomrule
    \end{tabular}
    \label{tab:main-results}
\end{table*}

\section{Results and Analysis}
\label{results}

\begin{table*}[ht]
\centering
\caption{Evaluation results on the test set for baseline Llama3 (Base) and finetuned PPACE models (Fine) in terms of Precision, Recall, and F-score for each individual category. Improvements (Imp) are also reported for these measures.}
\small
\begin{tabular}{@{}lcccccccccc@{}}
\toprule
\textbf{Category} & \textbf{Base P} & \textbf{Base R} & \textbf{Base F1} & \textbf{Fine P} & \textbf{Fine R} & \textbf{Fine F1} & \textbf{P Imp} & \textbf{R Imp} & \textbf{F1 Imp} \\ \midrule
Pathogen & 0.576 & 0.856 & 0.689 & 0.765 & 0.854 & 0.807 & 0.189 & -0.002 & 0.118 \\ 
Animal \& Dis. Vectors & 0.484 & 0.804 & 0.604 & 0.915 & 0.768 & 0.835 & 0.431 & -0.036 & 0.231 \\ 
Epidemiological & 0.565 & 0.476 & 0.517 & 0.721 & 0.646 & 0.682 & 0.156 & 0.171 & 0.165 \\ 
Clinical Char. in Humans & 0.399 & 0.808 & 0.535 & 0.775 & 0.603 & 0.678 & 0.376 & -0.205 & 0.144 \\ 
Infection Prev. \& Control & 0.150 & 0.696 & 0.247 & 0.677 & 0.457 & 0.545 & 0.527 & -0.239 & 0.298 \\ 
Therapeutics & 0.226 & 0.775 & 0.350 & 0.749 & 0.856 & 0.799 & 0.522 & 0.081 & 0.449 \\ 
Vaccines & 0.130 & 0.776 & 0.222 & 0.748 & 0.793 & 0.770 & 0.618 & 0.017 & 0.548 \\ 
Ethics & 0.133 & 0.200 & 0.160 & 1.000 & 0.150 & 0.261 & 0.867 & -0.050 & 0.101 \\ 
Public Health & 0.000 & 0.000 & 0.000 & 0.802 & 0.552 & 0.654 & 0.802 & 0.552 & 0.654 \\ 
Secondary Impacts & 0.391 & 0.391 & 0.391 & 0.777 & 0.904 & 0.836 & 0.386 & 0.513 & 0.445 \\ 
Health Systems & 0.133 & 0.730 & 0.225 & 0.429 & 0.169 & 0.242 & 0.296 & -0.562 & 0.017 \\ 
Capacity Strengthening & 0.015 & 0.600 & 0.028 & 0.000 & 0.000 & 0.000 & -0.015 & -0.600 & -0.028 \\ 
\bottomrule
\end{tabular}
\label{tab:results-test}
\end{table*}

Table \ref{tab:main-results} shows the results of the baseline Llama3-8B model, the larger Llama3-70b model, and the proposed PPACE model, respectively. As can be seen, PPACE outperforms the baselines on all the metrics with the exception of macro-averaged recall.

\begin{figure*}[ht]
    \centering
    \includegraphics[width=0.8\textwidth]{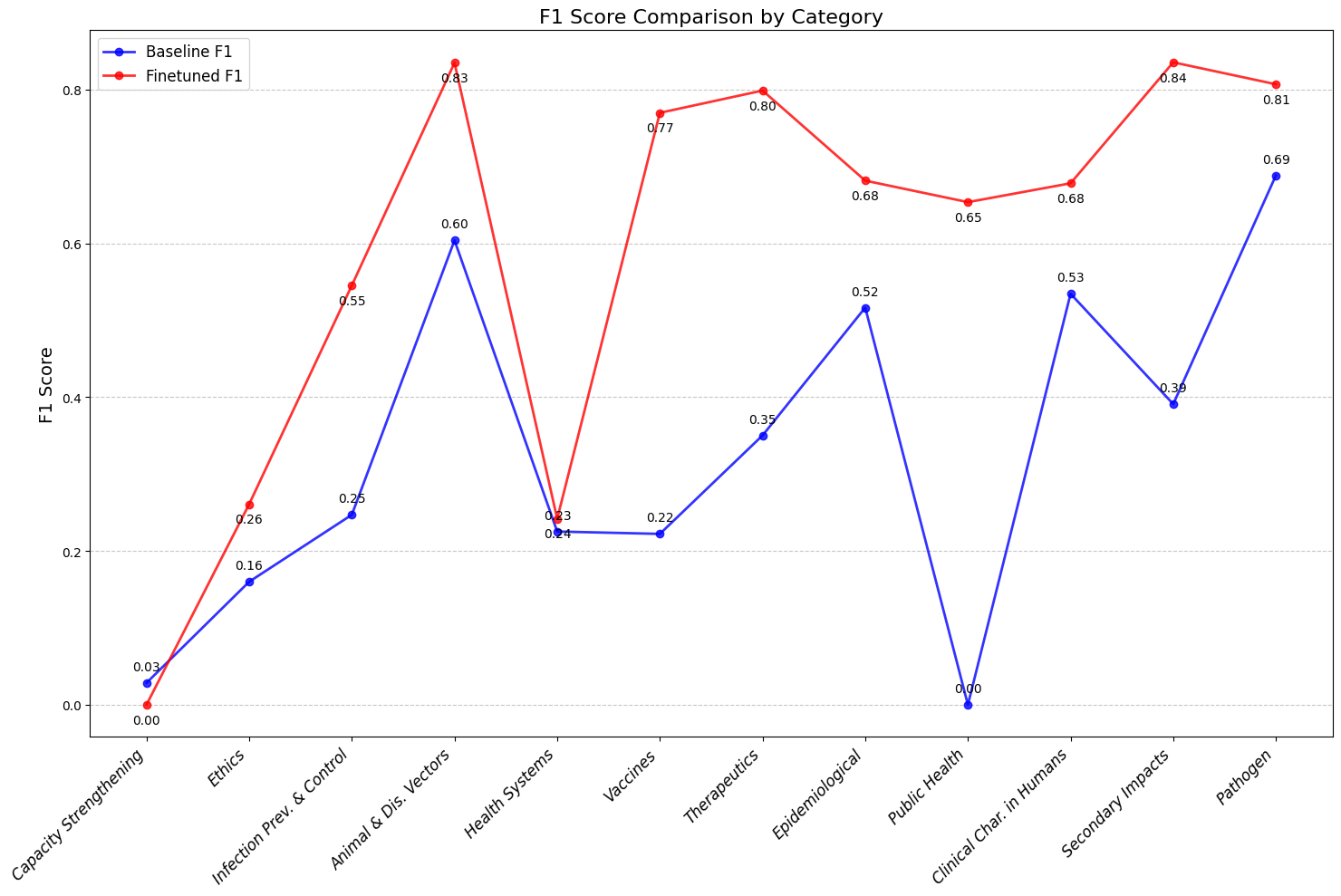}
    \caption{F1 Score Comparison by Category between the baseline Llama3 8B and the finetuned PPACE model. The categories are sorted from least to most frequent as seen in the test set.}
    \label{fig:fscore-changes}
\end{figure*}

The finetuned model demonstrates significant improvements in F1 scores across most categories, indicating the effectiveness of the finetuning process. Notable improvements are seen in categories like `Infection Prevention \& Control', `Therapeutics', and `Vaccines', where the finetuned model's precision and F1 scores show substantial gains. Categories with low representation in the dataset, such as `Capacity Strengthening' and `Health Systems', see mixed results, with some performance metrics slightly decreasing\footnote{The distribution of each category in the test set is provided in the Appendix Section \ref{sec:appendix_test_dist}}. This suggests that while finetuning enhances the model's ability to generalise, it may still struggle with categories that have very few examples. Figure \ref{fig:fscore-changes} shows the changes in F-scores between the fine-tuned and the base model across the different categories, sorted from the least frequent to the most frequent as seen in the test set.

Overall, the finetuned model generally performs better in terms of precision compared to recall. This trend indicates that the model has become more conservative in its predictions, leading to fewer false positives but potentially more false negatives. Categories like `Vaccines', `Public Health' and `Secondary Impacts' show remarkable improvements in precision and F1 scores, demonstrating PPACE's enhanced capability to identify relevant instances within these categories. The dramatic increase in all metrics for `Public Health' is particularly noteworthy, with the F-score jumping from 0 to 0.65. However, there remains room for improvement, especially for categories with minimal representation in the dataset. The results highlight the strengths of the finetuning approach while also pointing out the difficulty of the task.

\section{Conclusion}
\label{conclusion}

In this work we introduced the Pandemic PACT Advanced Categorisation Engine model or PPACE, a fine-tuned 8B language model for biomedical research classification as part of the Pandemic PACT initiative. PPACE is  capable of accurately categorising research abstracts according to WHO-aligned priorities. This tool can be a valuable asset for identifying biomedical research trends and gaps in a multilabel classification scenario. The model was built on a robust foundation of human-annotated data, enhanced with LLM-generated rationales, ensuring that the model's predictions are not only accurate but also interpretable. The use of efficient fine-tuning has enabled us to adapt the model effectively while maintaining computational efficiency.

Our evaluation demonstrated that PPACE outperforms its baselines, offering significant improvements in the context of multilabel classification. The model and the instruction-based dataset used for training are released oublicly, providing a valuable resource for the research community. These contributions facilitate further advancements in aligning biomedical research with critical global health priorities.

Looking ahead, the integration of LLMs in the annotation process promises to streamline data collection and categorisation, potentially reducing the burden on human annotators and improving the scalability of such initiatives. The evolution of PPACE can play a crucial role in enhancing the efficiency and effectiveness of global health research, ultimately contributing to better preparedness for future outbreaks.

\section*{Limitations}
Our work has several limitations. First, the dataset used for training and evaluation, while extensive, may not encompass the full diversity of biomedical research projects globally, potentially limiting the generalisability of our model for prospective analyses of research in new emerging pathogens. Additionally, the research categories might become outdated at some point, requiring updates and subsequent retraining of the model. Second, some projects can be categorised in different ways, introducing a degree of subjectivity in certain assignments. While our use of human-annotated expert labels aims to minimise this issue, it does not completely eliminate it.

Furthermore, despite using efficient fine-tuning methods like LoRA, the 8-billion parameter model is still sizable. Researchers with limited computational resources would need reliable GPUs for inference, as running solely on CPU can be very slow. Future iterations of this work will aim to fine-tune smaller models to improve accessibility.

Lastly, we did not experiment heavily with advanced prompting techniques or invest significant time in crafting the best possible prompts. There is potential for further improvement in the reported results for the frozen language models through optimised prompts, which might narrow the performance gap with the fine-tuned model.

\section*{Acknowledgments}
This research was funded by the National Institute for Health Research (NIHR) (CSA2022GloPID-R-3387) using UK Aid from the UK Government to support global health research, as part of the EDCTP2 Programme supported by the European Union. This research was also funded by the Wellcome Trust [226543/Z/22/Z]. For the purpose of open access, the author has applied a CC-BY public copyright licence to any author-accepted manuscript version arising from this submission. 

This work was also supported in part by the National Institute for Health Research (NIHR) Oxford Biomedical Research Centre (BRC), and in part by an InnoHK Project at the Hong Kong Centre for Cerebro-cardiovascular Health Engineering (COCHE). DAC was supported by an NIHR Research Professorship, an RAEng Research Chair, COCHE, the UKRI, and the Pandemic Sciences Institute at the University of Oxford. The views expressed are those of the authors and not necessarily those of the NIHR, COCHE, UKRI, NIHR, the UK Department of Health and Social Care, EDCTP, or the University of Oxford.

% Bibliography entries for the entire Anthology, followed by custom entries
%\bibliography{anthology,custom}
% Custom bibliography entries only
\bibliography{custom}

\begin{thebibliography}{23}
\providecommand{\natexlab}[1]{#1}

\bibitem[{Arighi et~al.(2011)Arighi, Roberts, Agarwal, Bhattacharya, Cesareni, Chatr-Aryamontri, Clematide, Gaudet, Giglio, Harrow et~al.}]{arighi2011biocreative}
Cecilia~N Arighi, Phoebe~M Roberts, Shashank Agarwal, Sanmitra Bhattacharya, Gianni Cesareni, Andrew Chatr-Aryamontri, Simon Clematide, Pascale Gaudet, Michelle~Gwinn Giglio, Ian Harrow, et~al. 2011.
\newblock Biocreative iii interactive task: an overview.
\newblock \emph{BMC bioinformatics}, 12:1--21.

\bibitem[{Baker et~al.(2015)Baker, Silins, Guo, Ali, H{\"o}gberg, Stenius, and Korhonen}]{baker2015automatic}
Simon Baker, Ilona Silins, Yufan Guo, Imran Ali, Johan H{\"o}gberg, Ulla Stenius, and Anna Korhonen. 2015.
\newblock Automatic semantic classification of scientific literature according to the hallmarks of cancer.
\newblock \emph{Bioinformatics}, 32(3):432--440.

\bibitem[{Behera et~al.(2019)Behera, Kumaravelan, and Kumar}]{behera2019performance}
Bichitrananda Behera, G~Kumaravelan, and Prem Kumar. 2019.
\newblock Performance evaluation of deep learning algorithms in biomedical document classification.
\newblock In \emph{2019 11th international conference on advanced computing (ICoAC)}, pages 220--224. IEEE.

\bibitem[{Carroll et~al.(2021)Carroll, Morzaria, Briand, Johnson, Morens, Sumption, Tomori, and Wacharphaueasadee}]{carroll2021preventing}
Dennis Carroll, Subhash Morzaria, Sylvie Briand, Christine~Kreuder Johnson, David Morens, Keith Sumption, Oyewale Tomori, and Supaporn Wacharphaueasadee. 2021.
\newblock Preventing the next pandemic: the power of a global viral surveillance network.
\newblock \emph{BMJ}, 372.

\bibitem[{Chen et~al.(2020)Chen, Allot, and Lu}]{chenlitcovid2020}
Qingyu Chen, Alexis Allot, and Zhiyong Lu. 2020.
\newblock \href {https://doi.org/10.1093/nar/gkaa952} {{LitCovid: an open database of COVID-19 literature}}.
\newblock \emph{Nucleic Acids Research}, 49(D1):D1534--D1540.

\bibitem[{Chen et~al.(2023)Chen, Du, Hu, Keloth, Peng, Raja, Zhang, Lu, and Xu}]{chen2023large}
Qingyu Chen, Jingcheng Du, Yan Hu, Vipina~Kuttichi Keloth, Xueqing Peng, Kalpana Raja, Rui Zhang, Zhiyong Lu, and Hua Xu. 2023.
\newblock Large language models in biomedical natural language processing: benchmarks, baselines, and recommendations.
\newblock \emph{arXiv preprint arXiv:2305.16326}.

\bibitem[{Dernoncourt and Lee(2017)}]{dernoncourt2017pubmed}
Franck Dernoncourt and Ji~Young Lee. 2017.
\newblock Pubmed 200k rct: a dataset for sequential sentence classification in medical abstracts.
\newblock In \emph{Proceedings of the Eighth International Joint Conference on Natural Language Processing (Volume 2: Short Papers)}, pages 308--313.

\bibitem[{Hersh et~al.(2006)Hersh, Cohen, Roberts, and Rekapalli}]{hersh2006trec}
William~R Hersh, Aaron~M Cohen, Phoebe~M Roberts, and Hari~Krishna Rekapalli. 2006.
\newblock Trec 2006 genomics track overview.
\newblock In \emph{TREC}, volume~7, pages 500--274.

\bibitem[{Hsieh et~al.(2023)Hsieh, Li, Yeh, Nakhost, Fujii, Ratner, Krishna, Lee, and Pfister}]{hsieh2023distilling}
Cheng-Yu Hsieh, Chun-Liang Li, Chih-kuan Yeh, Hootan Nakhost, Yasuhisa Fujii, Alex Ratner, Ranjay Krishna, Chen-Yu Lee, and Tomas Pfister. 2023.
\newblock Distilling step-by-step! outperforming larger language models with less training data and smaller model sizes.
\newblock In \emph{Findings of the Association for Computational Linguistics: ACL 2023}, pages 8003--8017.

\bibitem[{Hu et~al.(2021)Hu, Shen, Wallis, Allen-Zhu, Li, Wang, Wang, and Chen}]{hu2021lora}
Edward~J Hu, Yelong Shen, Phillip Wallis, Zeyuan Allen-Zhu, Yuanzhi Li, Shean Wang, Lu~Wang, and Weizhu Chen. 2021.
\newblock Lora: Low-rank adaptation of large language models.
\newblock \emph{arXiv preprint arXiv:2106.09685}.

\bibitem[{Jiang et~al.(2024)Jiang, Huang, Luo, Zhang, Huang, Wei, Deng, Sun, Zhang, Wang et~al.}]{jiang2024mora}
Ting Jiang, Shaohan Huang, Shengyue Luo, Zihan Zhang, Haizhen Huang, Furu Wei, Weiwei Deng, Feng Sun, Qi~Zhang, Deqing Wang, et~al. 2024.
\newblock Mora: High-rank updating for parameter-efficient fine-tuning.
\newblock \emph{arXiv preprint arXiv:2405.12130}.

\bibitem[{Jimenez~Gutierrez et~al.(2020)Jimenez~Gutierrez, Zeng, Zhang, Zhang, and Su}]{jimenez-gutierrez-etal-2020-document}
Bernal Jimenez~Gutierrez, Jucheng Zeng, Dongdong Zhang, Ping Zhang, and Yu~Su. 2020.
\newblock \href {https://doi.org/10.18653/v1/2020.findings-emnlp.332} {Document classification for {COVID}-19 literature}.
\newblock In \emph{Findings of the Association for Computational Linguistics: EMNLP 2020}, pages 3715--3722, Online. Association for Computational Linguistics.

\bibitem[{Laza et~al.(2011)Laza, Pav{\'o}n, Reboiro-Jato, and Fdez-Riverola}]{laza2011evaluating}
Rosal{\'\i}a Laza, Reyes Pav{\'o}n, Miguel Reboiro-Jato, and Florentino Fdez-Riverola. 2011.
\newblock Evaluating the effect of unbalanced data in biomedical document classification.
\newblock \emph{Journal of integrative bioinformatics}, 8(3):105--117.

\bibitem[{Luo and Specia(2024)}]{luo2024understanding}
Haoyan Luo and Lucia Specia. 2024.
\newblock From understanding to utilization: A survey on explainability for large language models.
\newblock \emph{arXiv preprint arXiv:2401.12874}.

\bibitem[{McCoy et~al.(2023)McCoy, Yao, Friedman, Hardy, and Griffiths}]{mccoy2023embers}
R~Thomas McCoy, Shunyu Yao, Dan Friedman, Matthew Hardy, and Thomas~L Griffiths. 2023.
\newblock Embers of autoregression: Understanding large language models through the problem they are trained to solve.
\newblock \emph{arXiv preprint arXiv:2309.13638}.

\bibitem[{McLean et~al.(2022)McLean, Rashan, Tran, Arena, Lawal, Maguire, Adele, Antonio, Brack, Caldwell et~al.}]{mclean2022fragmented}
Alistair~RD McLean, Sumayyah Rashan, Lien Tran, Lorenzo Arena, AbdulAzeez Lawal, Brittany~J Maguire, Sandra Adele, Emilia~Sitsofe Antonio, Matthew Brack, Fiona Caldwell, et~al. 2022.
\newblock The fragmented covid-19 therapeutics research landscape: a living systematic review of clinical trial registrations evaluating priority pharmacological interventions.
\newblock \emph{Wellcome Open Research}, 7(24):24.

\bibitem[{Norton et~al.(2020)Norton, Sigfrid, Aderoba, Nasir, Bannister, Collinson, Lee, Boily-Larouche, Golding, Depoortere et~al.}]{norton2020preparing}
Alice Norton, Louise Sigfrid, Adeniyi Aderoba, Naima Nasir, Peter~G Bannister, Shelui Collinson, James Lee, Genevi{\`e}ve Boily-Larouche, Josephine~P Golding, Evelyn Depoortere, et~al. 2020.
\newblock Preparing for a pandemic: highlighting themes for research funding and practice—perspectives from the global research collaboration for infectious disease preparedness (glopid-r).
\newblock \emph{BMC medicine}, 18:1--4.

\bibitem[{Norton et~al.(2024)Norton, Sigfrid, Antonio, Bucher, Ndwandwe, and Group}]{norton2024improving}
Alice Norton, Louise Sigfrid, Emilia Antonio, Adrian Bucher, Duduzile Ndwandwe, and Pandemic PACT~Advisory Group. 2024.
\newblock Improving coherence of global research funding: Pandemic pact.
\newblock \emph{Lancet (London, England)}, 403(10433):1233.

\bibitem[{Phan et~al.(2021)Phan, Anibal, Tran, Chanana, Bahadroglu, Peltekian, and Altan-Bonnet}]{phan2021scifive}
Long~N Phan, James~T Anibal, Hieu Tran, Shaurya Chanana, Erol Bahadroglu, Alec Peltekian, and Gr{\'e}goire Altan-Bonnet. 2021.
\newblock Scifive: a text-to-text transformer model for biomedical literature.
\newblock \emph{arXiv preprint arXiv:2106.03598}.

\bibitem[{Rohanian et~al.(2023)Rohanian, Nouriborji, and Clifton}]{rohanian2023exploring}
Omid Rohanian, Mohammadmahdi Nouriborji, and David~A Clifton. 2023.
\newblock Exploring the effectiveness of instruction tuning in biomedical language processing.
\newblock \emph{arXiv preprint arXiv:2401.00579}.

\bibitem[{Seminog et~al.(2024)Seminog, Furst, Mendy et~al.}]{Seminog2024}
O~Seminog, R~Furst, T~Mendy, et~al. 2024.
\newblock \href {https://doi.org/10.12688/wellcomeopenres.21202.1} {A protocol for a living mapping review of global research funding for infectious diseases with a pandemic potential – pandemic pact}.
\newblock \emph{Wellcome Open Research}, 9:156.
\newblock Version 1; peer review: awaiting peer review.

\bibitem[{Taylor et~al.(2024)Taylor, Ghose, Rohanian, Nouriborji, Kormilitzin, Clifton, and Nevado-Holgado}]{taylor2024efficiency}
Niall Taylor, Upamanyu Ghose, Omid Rohanian, Mohammadmahdi Nouriborji, Andrey Kormilitzin, David Clifton, and Alejo Nevado-Holgado. 2024.
\newblock Efficiency at scale: Investigating the performance of diminutive language models in clinical tasks.
\newblock \emph{arXiv preprint arXiv:2402.10597}.

\bibitem[{Tian et~al.(2024)Tian, Jin, Yeganova, Lai, Zhu, Chen, Yang, Chen, Kim, Comeau et~al.}]{tian2024opportunities}
Shubo Tian, Qiao Jin, Lana Yeganova, Po-Ting Lai, Qingqing Zhu, Xiuying Chen, Yifan Yang, Qingyu Chen, Won Kim, Donald~C Comeau, et~al. 2024.
\newblock Opportunities and challenges for chatgpt and large language models in biomedicine and health.
\newblock \emph{Briefings in Bioinformatics}, 25(1):bbad493.

\end{thebibliography}

\appendix
\section*{Appendix}
\section{List of Research Categories Along with Definitions}
\label{sec:appendix}

The following is the list of categories used in the prompt along with corresponding explanations of what each category entails: 

\begin{enumerate}
    \item \textbf{Pathogen: Natural History, Transmission, and Diagnostics:} Development of diagnostic tools, understanding pathogen morphology, genomics, and genotyping, studying immunity, using disease models, and assessing the environmental stability of pathogens.
    
    \item \textbf{Animal and Environmental Research \& Research on Diseases Vectors:} Animal sources, transmission routes, vector biology, and control strategies for vectors.
    
    \item \textbf{Epidemiological Studies:} Research on disease transmission dynamics, susceptibility, control measure effectiveness, and disease mapping through surveillance and reporting.
    
    \item \textbf{Clinical Characterisation and Management in Humans:} Prognostic factors for disease severity, disease pathogenesis, supportive care and management, long-term health consequences, and clinical trials for disease management.
    
    \item \textbf{Infection Prevention and Control:} Research on community restriction measures, barriers and PPE, infection control in healthcare settings, and measures at the human-animal interface.
    
    \item \textbf{Therapeutics Research, Development, and Implementation:} Pre-clinical studies for therapeutic development, clinical trials for therapeutic safety and efficacy, development of prophylactic treatments, logistics and supply chain management for therapeutics, clinical trial design for therapeutics, and research on adverse events related to therapeutic administration.
    
    \item \textbf{Vaccines Research, Development, and Implementation:} Pre-clinical studies for vaccine development, clinical trials for vaccine safety and efficacy, logistics and distribution strategies for vaccines, vaccine design and administration, clinical trial design for vaccines, research on adverse events related to immunisation, and characterisation of vaccine-induced immunity.
    
    \item \textbf{Research to Inform Ethical Issues:} Ethical considerations in research design, ethical issues in public health measures, ethical clinical decision-making, ethical resource allocation, ethical governance, and ethical considerations in social determinants of health.
    
    \item \textbf{Policies for Public Health, Disease Control and Community Resilience:} Approaches to public health interventions, community engagement, communication and infodemic management, vaccine/therapeutic hesitancy, and policy research and interventions.
    
    \item \textbf{Secondary Impacts of Disease, Response, and Control Measures:} Indirect health impacts, social impacts, economic impacts, and other secondary impacts such as environmental effects, food security, and infrastructure.
    
    \item \textbf{Health Systems Research:} Health service delivery, health financing, access to medicines and technologies, health information systems, health leadership and governance, and health workforce management.
    
    \item \textbf{Capacity Strengthening:} Individual capacity building, institutional capacity strengthening, systemic/environmental components, and cross-cutting activities across all levels of capacity building.
\end{enumerate}

\section{Prompt Design for Model Inference}
\label{sec:appendix_prompt}

To generate high-quality inferences from the models during the few-shot learning experiments, a carefully constructed prompt was employed. This prompt was designed based on feedback from expert human annotators to ensure precision and discourage spurious associations. Below is the detailed explanation and the code used for generating the prompt.

The prompt includes guidelines and examples to steer the model towards accurate classification. The guidelines ensure that the model focuses on relevant categories and avoids unnecessary implications or speculative guesses. The examples provided demonstrate the expected structure and reasoning for categorisation.

The \texttt{\{guideline\}} variable in the prompt is replaced with the list of research categories mentioned at \ref{sec:appendix}, prefaced with the following line:

\texttt{We have a project in the area of biomedical research which we want to classify in terms of the research priorities it relates to. We have 12 possible research priorities and a project can be mapped to 1 or more of these priorities. The following is a guide on what each of these 12 categories are alongside the specific areas that they cover.}

\begin{lstlisting}[language=Python, caption=Function to generate the prompt for classifying research projects]
# Function to generate the prompt for classifying research projects
def generate_classification_prompt(row, guideline):
    # Assume row is a pandas Series with project info and guideline is a string containing the guidelines.

    print(guideline)

    prompt = f"""
    [INST] Based on the research categorization guidelines, classify the following project into the appropriate primary research priorities using only the top-level categories 1 to 12. Structure your response clearly, providing the category numbers enclosed in single quotation marks.

    {guideline}

    Examples:
    - For a study on investigating the genetic mutations of a pathogen and its resistance to current vaccines.
      ### Reasoning: Categories '1' and '7' are chosen for their focus on pathogen genomics and vaccine development, respectively.
      ### Categories: '1', '7'

    - For a study on examining the effectiveness of new therapeutic treatments in Phase 3 clinical trials and the ethical considerations in conducting these trials.
      ### Reasoning: Categories '6' and '8' are chosen for their focus on Phase 3 clinical trials and ethical research issues, respectively.
      ### Categories: '6', '8'

    - For a study on the social determinants of disease spread in urban environments, the efficacy of non-pharmaceutical interventions, and the long-term mental health impacts on survivors.
      ### Reasoning: Categories '3', '9', and '10' are selected for their relevance to disease transmission, public health interventions, and indirect health impacts.
      ### Categories: '3', '9', '10'

    Note 1: Use category '2' only for explicit references to animals (this is a rare category).
    Note 2: Research Collaboration is distinct from epidemiological studies.
    Note 3: Don't categorize solely on study population.
    Note 4: Therapeutics Research pertains to drugs.
    Note 5: Stay logical and factual in your analysis. Avoid making unnecessary implications or speculative guesses beyond the explicit information provided.

    Based on this information, identify the relevant research categories for this project. Provide clear but succinct reasoning for your choices similar to the above examples. Section your response in the following format:

    ### Reasoning: ...
    ### Categories: ...

    Project Information:
    Title: {row['Grant Title Eng']}
    Abstract: {row['Abstract Eng']}
    [/INST]""".strip()

    return prompt
\end{lstlisting}

In the case of fine-tuning, in order to ensure the model is actually learning from the labels rather than relying on extra information in the prompt, we do not use the few-shot examples and omit the extra 5 notes as well. Every other detail in the template above stays the same when finetuning.  

\section{Distribution of Categories in the Test set}
\label{sec:appendix_test_dist}

\begin{figure}[ht]
\centering
\includegraphics[width=0.95\columnwidth]{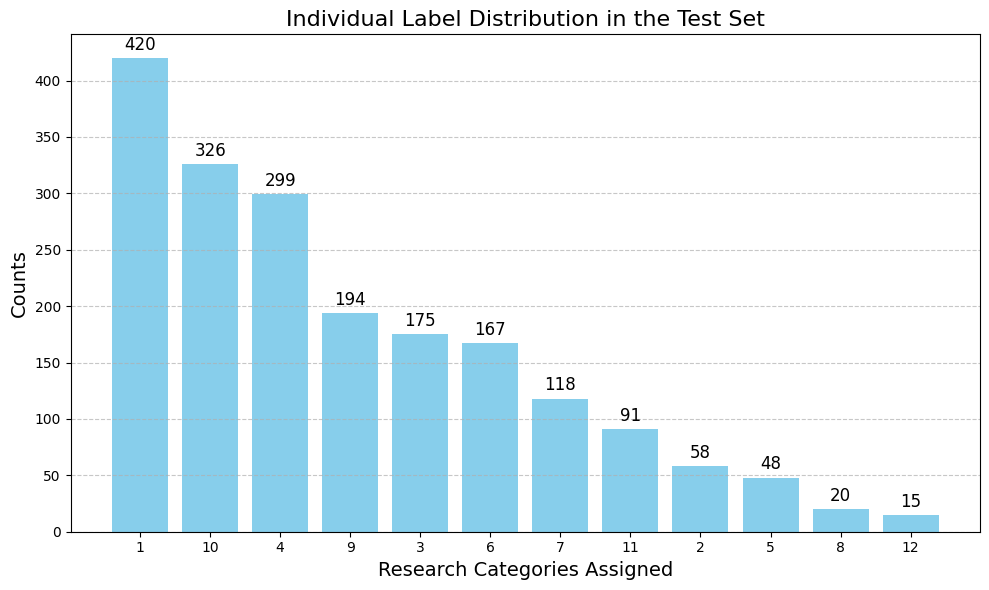}
\caption{Individual Label Distribution in the Test Set.}
\label{fig:individual-label-dist-test}
\end{figure}

\end{document}